# Comparison of Two Methods for Stationary Incident Detection Based on Background Image

Deepak Ghimire (deep@jbnu.ac.kr), Joonwhoan Lee (chlee@jbnu.ac.kr)

**Abstract**

In general, background subtraction-based methods are used to detect moving objects in visual tracking applications. In this paper, we employed a background subtraction-based scheme to detect temporarily stationary objects. We proposed two schemes for stationary object detection and compared them in terms of detection performance and computational complexity. In the first approach, we used a single background, and in the second approach, we used dual backgrounds generated with different learning rates to detect temporarily stopped objects. Finally, we used normalized cross-correlation (NCC)-based image comparison to monitor and track the detected stationary objects in a video scene. The proposed method is robust against partial occlusion, short-term full occlusion, and illumination changes, and it can operate in real time.

Keywords: Background Subtraction | Temporarily Stationary Objects | Normalized Cross Correlation | Object Tracking | Illegally Parked Vehicles

## I. Introduction

Temporarily static object detection has many applications. Depending on the application, temporarily static objects can be abandoned items such as luggage, illegally parked vehicles, removed objects from the scene, etc. A significant amount of research has been done on the detection of abandoned items [1, 2, 3] in video surveillance, illegally parked vehicle detection [4, 5, 6, 7, 8, 9] in restricted regions, and detection of removed objects from the scene [10]. In [1], they first used a trans-dimensional Markov Chain Monte Carlo tracking model to track objects in a scene. Then, the result of the tracking system is analyzed, and left luggage is detected in the scene. Background subtraction and blob tracking are used in [2] to detect abandoned items. Short-term logic is applied to classify the detected blobs into four types: unknown objects, abandoned objects, person, and still person. Similarly, in [3], background subtraction and tracking are used to detect left luggage using multiple cameras. Most of these methods assume the scene is not crowded, with no occlusion or illumination changes.

Bevilacqua and Vaccari [4] proposed a method to detect stopped vehicles based on the centroid position of the tracked vehicle. Background subtraction and optical flow methods are used for detection and tracking of stopped vehicles.

\* This work was partially supported by National Research Foundation of Korea (2011-0022152)

Manuscript: 2012. 09. 23  
Revised: 2012. 09. 08

Corresponding Author: Joonwhoan Lee  
e-mail: chlee@jbnu.ac.kr

If the object's center position remains within a small area for a certain duration, the object is considered static. A temporarily static object detection method based on two backgrounds—one short-term and another long-term—is presented in [5, 6]. Object tracking-based methods are also used for detecting static objects in a scene. In [7], J. T. Lee et al. presented a 1-D transformation-based real-time illegal parking detection method. They first apply a 1-D transformation to the source video data. Next, foreground blobs representing vehicles are segmented and tracked frame by frame. Parking vehicles are detected according to the trajectory of the tracking result.

B. Mitra et al. [8] presented an illegally parked vehicle tracking method using correlation of multi-scale difference of Gaussian filtered patches. Similarly, in [9], a corner feature-based parked vehicle detection method is presented. They classified corners into two categories: static and dynamic. Dynamic corners correspond to moving objects, and static corners correspond to the background and stopped objects. A disadvantage of this method is that static corners corresponding to the background can be mistakenly detected as corners corresponding to stopped vehicles, and vice versa. Finally, in [10], abandoned and removed object detection based on background subtraction and foreground analysis complemented by tracking is presented. Most of the methods above work well if objects are static for a short duration. However, if temporarily static objects remain in the scene for a long time, they may appear as background objects. Therefore, an additional process is needed to monitor static objects as soon as they are detected as abandoned or stopped.

In this paper, we present and compare two methods for detecting temporarily static objects. We particularly focus on illegally parked vehicle detection, but the same methods can be applied to detect any kind of temporarily static object in a scene. We used the Gaussian Mixture Model (GMM) method for separating foreground objects from the background image. In the first method, we used background subtraction and a simple blob tracking method to classify objects as static or moving. As soon as an object is detected as static, we used normalized cross-correlation (NCC)-based image comparison to monitor the temporarily stopped object. In the second method, we used two background images generated using different learning rates or frame rates. We subtracted the two background images to detect static objects. Again, NCC-based image comparison is used to monitor the detected static objects in the scene. The main advantage of the proposed methods is that they can detect static objects in a scene even if they remain there for a long duration.

## II. Stationary Object Detection

In the first stage, we detect stationary objects in a scene. Once an object is identified as stationary, we proceed to the second stage for further processing. Here, we focus on describing the first stage. We propose two approaches for this stage:

using a single background image and using dual background images. Each approach is described in the following subsections.

## 1. Single Background Based Stationary Object Detection

In general, background subtraction is used to detect and track moving objects in a scene. If an object appears in the scene and becomes stationary, it will be detected as a foreground object for a while; over time, it will be incorporated into the background image and then classified as a background object. Therefore, there is a time interval during which we can decide whether a detected object by background subtraction is stationary or moving. Here, we use background subtraction followed by simple rectangle-overlapping based binary blob tracking to determine whether an object is stationary or moving. Fig. 1 shows the overall block diagram of the stationary object detection. It includes an input ROI image of a video frame, the corresponding background image, and the resulting foreground image detected using background subtraction and thresholding.

In Fig. 2(a), the upper vehicle is moving while the lower vehicle is stationary. Before starting the tracking, we apply frame differencing to reduce noise generated by moving objects. Fig. 3(a) shows the result of the frame difference, and Fig. 3(b) displays two types of pixels corresponding to foreground objects. Pixels labeled in gray correspond to moving object pixels (white pixels in Fig. 3(a)), whereas pixels labeled in white correspond to stationary foreground object pixels. Since we are interested in finding stationary objects, we remove the pixels corresponding to moving objects. The final result is shown in Fig. 3(d).

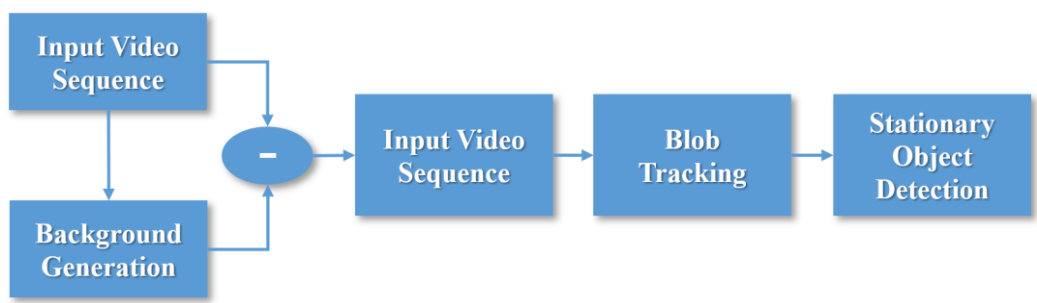

Fig. 1. Proposed system of stationary object detection using single background image.

Vehicle tracking is performed using a simple rectangle overlapping approach. For every frame, we label the binary blobs resulting from background subtraction and moving object pixel removal. If the bounding rectangle of a binary blob in the previous frame and the current frame overlap by more than 80%, we consider that bounding rectangle to correspond to the same object in both frames. If the object is detected as the same in several consecutive frames (greater than a predefined monitoring threshold), it is classified as a stationary object and moves to the second stage: stationary object monitoring.

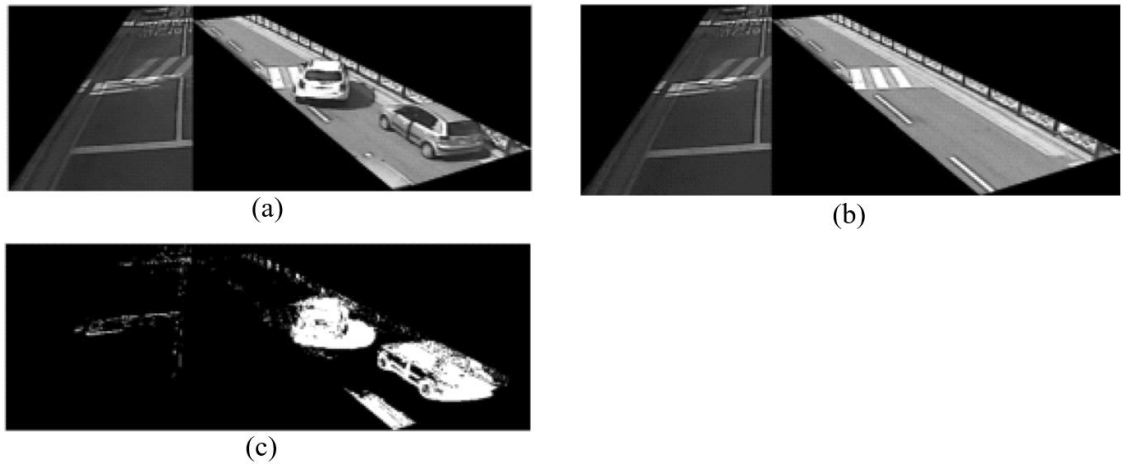

Fig. 2. **(a)** Current input image, **(b)** background image, **(c)** result of background subtraction and thresholding.

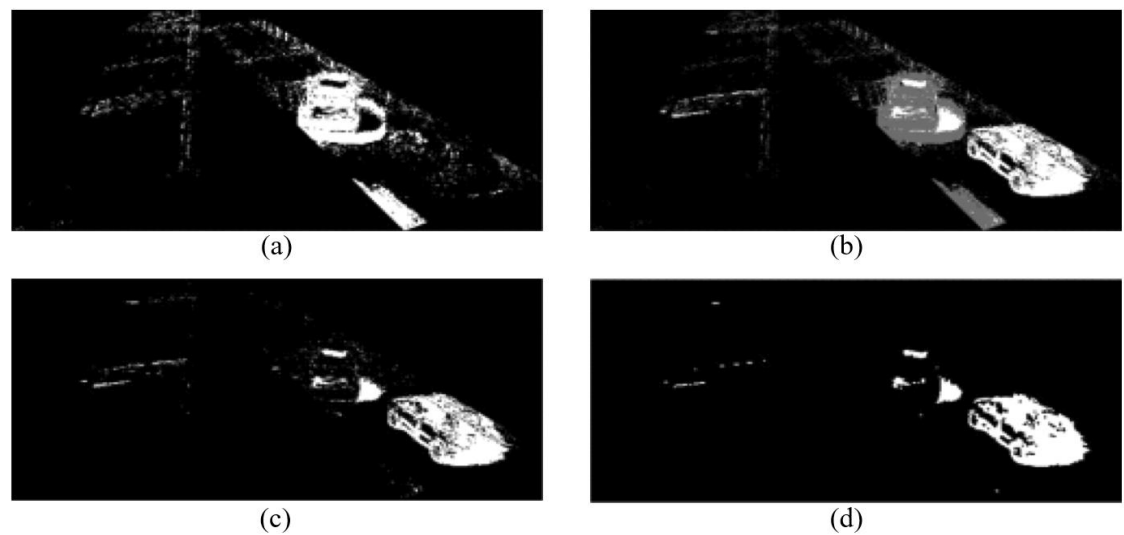

Fig. 3. **(a)** Frame difference image ($i_t - i_{t-1}$), **(b)** foreground image with gray pixels corresponding to moving objects (white pixels from (a)) and white pixels corresponding to stopped objects, **(c)** result after removing moving object pixels from (a), and **(d)** result after morphological erosion and dilation.

### 2. Dual Background Based Stationary Object Detection

Two background models, generated at different frame rates or learning rates, can be utilized to detect stationary objects. If an object remains stationary within the region of interest, it will first appear in the background image with a fast update rate, and after some time, it will also appear in the background image with a slow update rate. Thus, there is a time interval during which the stopped vehicle is visible in one background but not in the other. There are two approaches that can be used to generate fast and slow updating background images: one by processing at different frame rates, and the other by processing with different learning rates. Fig. 4 shows the block diagram of the proposed stationary object detection method using dual background images.

Fig. 5 displays the current input image (I), the fast-updating background (BGF), the slow-updating background (BGS), and the binary difference image between the two backgrounds (BGDIFF), which is the result after applying morphological erosion and dilation operations. As shown, there are several vehicles in the current image. The vehicles that appear in BGF are currently stopped vehicles, while others are moving vehicles. However, the stopped vehicles have not yet appeared in the BGS image. Therefore, by using the BGF and BGS images, we can compute the BGDIFF image, in which the binary blobs indicate the positions of stopped vehicles. Once a vehicle is detected as stopped across several frames, we move to the next stage, where the stopped object is monitored using the NCC method.

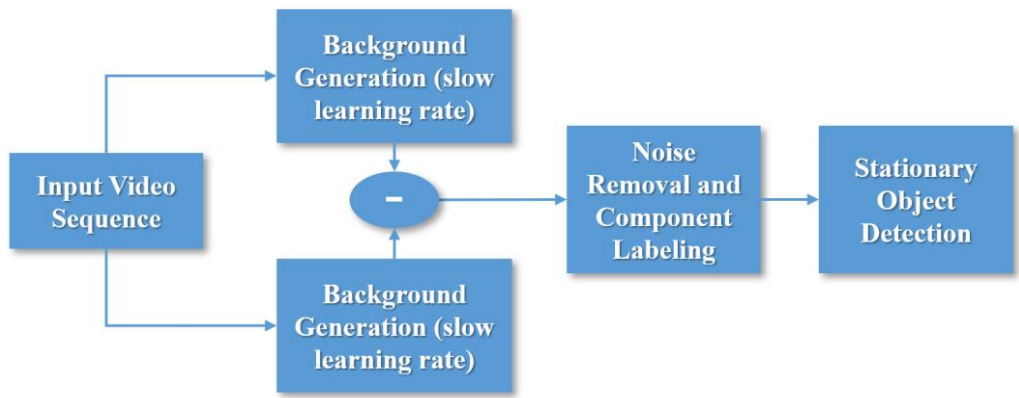

Fig. 4. Proposed system of stationary object detection using dual backgrounds.

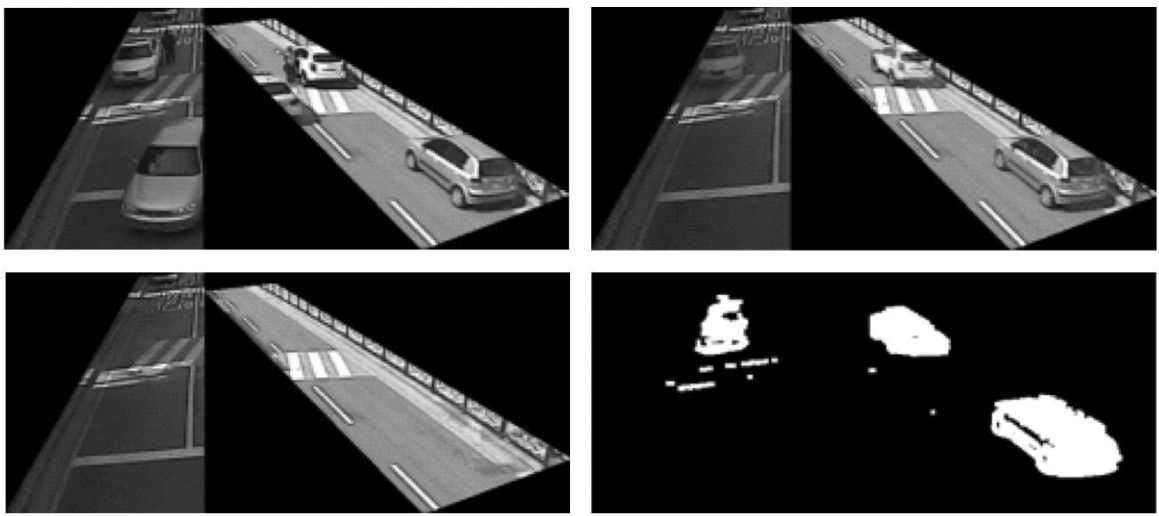

Fig. 5. Current image I (left top), background image with fast learning rate ($BG_F$) (right top), background image with slow learning rate ($BG_S$) (bottom left), and difference between $BG_F$ and $BG_S$ with thresholding and morphological operation (bottom right).

## III. Normalized Cross-Correlation Based Stationary Object Monitoring

Normalized Cross-Correlation (NCC) can be used to compare two signals to evaluate their similarity. The closer the NCC value is to 1, the more similar the two signals are. Therefore, NCC can be used to compare two images to determine how similar they are. In this work, we use the NCC value between two images to determine whether a temporarily stationary object is still at its detected position or has been removed or moved.

As soon as an object is detected as stationary, we register the image patch within the bounding rectangle of the corresponding stationary object as a reference image. For subsequent frames, the image patch from the same position in the current image is compared with the previously stored reference image patch. As long as the object remains at the same position, the NCC value remains close to 1. If the object is removed or moved, the NCC value drops significantly.

When there are multiple stationary objects, calculating NCC for each object in every frame becomes computationally expensive. To reduce computational complexity, NCC comparisons are performed approximately twice per second.

In another scenario, if there is occlusion (partial or full) caused by moving objects near the stationary object, the NCC value between the current image patch and the reference image patch may be low even if the stationary object is still present. To handle this issue, we first count the number of moving object pixels around the stationary object using a frame difference image. If the count exceeds a predefined threshold, the NCC comparison is postponed to the next frame.

Additionally, gradual changes in illumination over time can cause the calculated NCC value to decrease, even if the object has not moved or been removed. To address this, for objects that remain in place for a long duration, we periodically update the reference image patch after a certain interval.

The NCC value between the reference image patch and the current image patch is computed using Equation (1).

$$\gamma = \frac{\sum_{x,y}(f_r(x,y) - \bar{f}_r)(f_c(x,y) - \bar{f}_c)}{\sqrt{\sum_{x,y}(f_r(x,y) - \bar{f}_r)^2}\sqrt{\sum_{x,y}(f_c(x,y) - \bar{f}_c)^2}} \quad (1)$$

In $f_r$ denotes the reference patch image, $\bar{f}_r$ denotes the mean value of the reference patch image, $f_c$ denotes the current patch image, $\bar{f}_c$ denotes the mean value of the current patch image, and $\gamma$ denotes the resulting NCC value.

## IV. Experimental Results and Discussion

The performance of the proposed method is evaluated on a private dataset of surveillance video. The experiment focuses on detecting illegally parked vehicles. As soon as a vehicle enters the region under analysis, it is detected immediately if it stops there. The result of tracking or stopped vehicle detection at the first stage gives the time duration, or the number of frames, for which a particular vehicle remains stopped in the region under analysis. If it stops for more than 50 image frames, it is defined as a stopped vehicle. This means it will remain in the first stage for 50 to 150 frames.

If the vehicle moves out before 150 frames, it is classified as a stopped vehicle but not a parked vehicle. However, if the vehicle remains in the same position for more than 150 frames, it is defined as a parked vehicle, i.e., it enters the second stage.

When a stopped vehicle enters the second stage (parked stage), we use NCC (Normalized Cross-Correlation) to further verify that the vehicle is still parked, as long as it remains in place. In our case, if the NCC value between the reference image patch of the stopped vehicle's position and the current image patch at the same position is greater than or equal to 0.90, it indicates that the vehicle is still parked at the same position. This threshold value was determined experimentally. If the NCC value is less than 0.90, it indicates that the vehicle has moved, i.e., the third stage (moved stage).

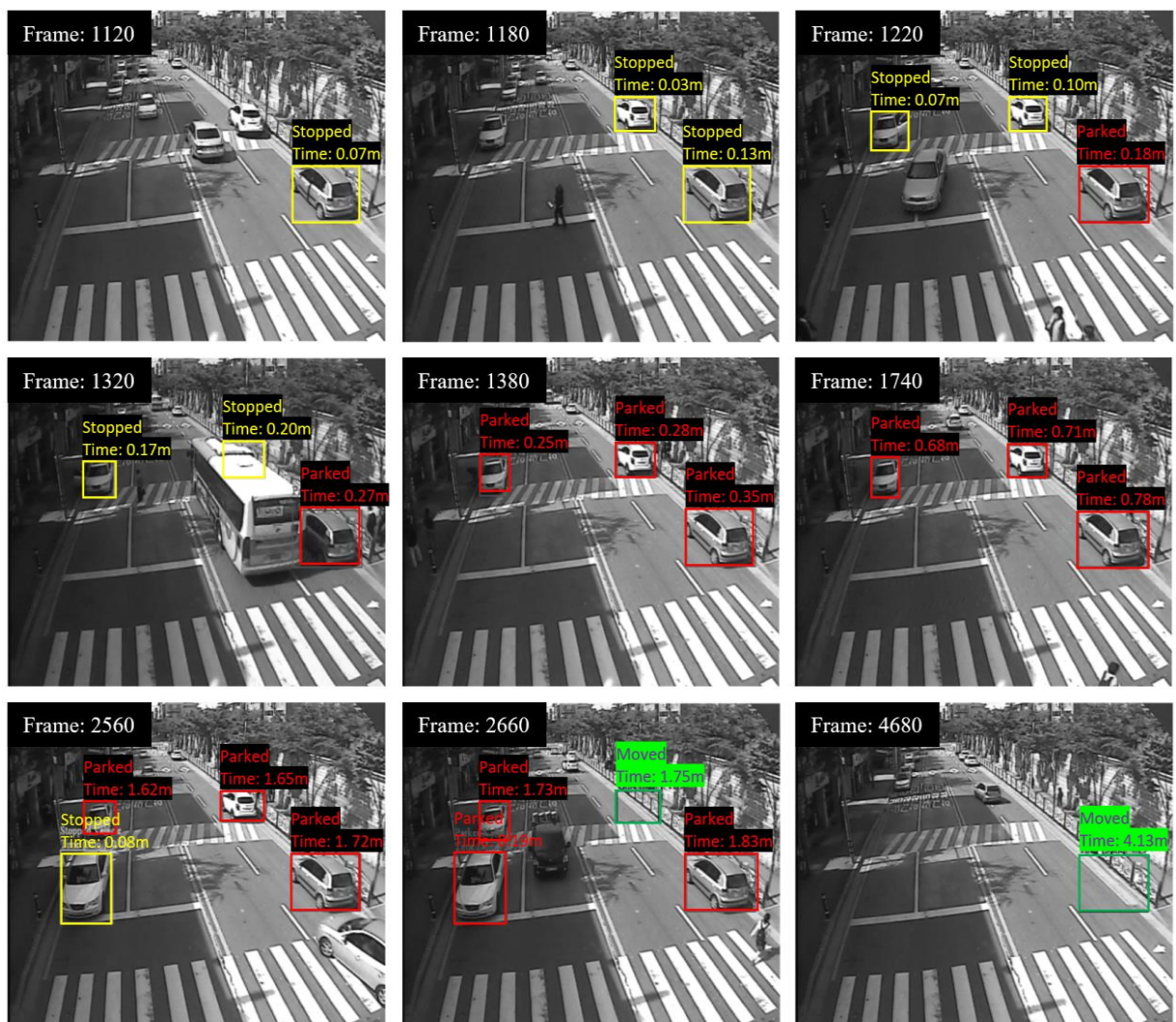

Fig. 6. Illegal parking vehivle detection results for a video sequence using dual background modeling scheme.

Figure 6 shows the result of illegal parking detection, with the time duration and stages provided for each stopped or parked vehicle. The given time duration represents the total time starting from when the vehicle stopped in the region under analysis. From Figure 6, we can see that the parking time of each vehicle has also been recorded. In this particular video, the maximum parking time was found to be 4.13 minutes. Additionally, there were no issues with partial occlusion or short-term full occlusion.

In our video dataset, the performance of the dual background-based stationary object detection is better than that of the single background-based method. In terms of computational complexity, the single background-based scheme is better and works well if there are no illumination changes, the surveillance scene is not crowded, and the stationary object is not frequently occluded by moving objects. On the other hand, the dual background-based scheme has higher computational complexity compared to the single-background-based method, but it can still operate in real time. Moreover, the dual-background-based scheme is robust to partial occlusion and short-term full occlusion.

Since we subtract two backgrounds to detect stationary objects, temporary occlusion does not significantly affect the background model. However, problems may arise if an object is occluded for a long duration. The dual background-based scheme is also robust to illumination changes. When subtracting a background from the current image, noise may occur due to moving objects, sudden illumination changes, or object shadows. However, subtracting two backgrounds generated with different learning rates captures only stationary objects, making the result more stable.

Table 1 shows the operating speed of the two proposed schemes for temporarily static object detection. Originally, the video frame size is 720 × 480 pixels. We extracted the regions of interest (ROI) from both sides of the street and concatenated them, as shown in the figures in Section 2. According to the results in Table 1, for the single-background case with an image size of 329 × 164 pixels, we achieved an operating speed of up to 33.26 frames per second. In contrast, for the dual-background scenario, we achieved a speed of up to 20.13 frames per second. Although the dual-background method has a lower operating speed, it is more stable than the single background-based scenario.

Table 1. Comparison of two proposed scheme of stationary object detection in terms of computation complexity.

| Scenario | Input Image (ROI) size (in pixels) | Processing Speed (in frames/second) |
|---|---|---|
| Single Background | 658 x 329 | 23.20 |
| Single Background | 329 x 164 | 33.26 |
| Dual Background | 658 x 329 | 8.18 |
| Dual Background | 329 x 158 | 20.13 |

## V. Conclusion

In this paper, we presented a comparative study of two scenarios for temporarily stationary object detection: one using a single background model and the other using dual background models. In both approaches, once an object is detected and classified as stationary, it is further monitored and tracked using NCC-based image comparison. The dual background-based method demonstrated superior and more stable performance compared to the single background-based method. Although the single background approach is computationally more efficient, it requires additional overhead for distinguishing between moving and static objects through tracking. We evaluated the proposed system on a surveillance video dataset focused on detecting illegally parked vehicles in restricted areas of a street. The results indicate that the system is robust and capable of operating in real time.

## References

[1] K. Smith, P. Quelhas, and D. Gatica-Perez, "Detecting abandoned luggage items in public spaces," *PETS*, pp. 75-82, 2006.

[2] N. Bird, S. Atev, N. Caremelli, R. Martin, O. Masoud, and N. Papanikolopoulos, "Real-time, online detection of abandoned objects in public areas," *IEEE Int. Conf. on Robotics and Automation*, pp. 3775-3780, 2006.

[3] J. M. del Rincón, J. E. Herrero-Jaraba, J. R. Gómez, and C. Orrite-Uruñuela, "Automatic left luggage detection and tracking using multi-camera UKF," *PETS*, pp. 59-66, 2006.

[4] A. Bevilacqua and S. Vaccari, "Real-time detection of stopped vehicles in traffic scenes," *Proc. IEEE Conf. on Advanced Video and Signal Based Surveillance*, pp. 266-270, 2007.

[5] F. Porikli, "Detection of temporarily static regions by processing video at different frame rates," *Proc. IEEE Conf. on Advanced Video and Signal Based Surveillance*, pp. 236-241, 2007.

[6] T. Y. Ping and C. Y. Yu, "Illegally parked vehicle detection based on omnidirectional computer vision," *Proc. IEEE Conf. on Image and Signal Processing*, pp. 1-5, 2009.

[7] J. T. Lee, M. S. Ryoo, M. Riley, and J. K. Aggarwal, "Real-time illegal parking detection in outdoor environments using 1-D transformation," *IEEE Trans. on Circuits and Systems for Video Technology*, vol. 19, no. 9, pp. 1014-1024, 2009.

[8] B. Mitra, W. Hassan, N. Bangalore, P. Brich, R. Young, and C. Chatwin, "Tracking illegally parked vehicles using correlation of multi-scale difference of Gaussian filtered patches," *Proc. SPIE*, vol. 8005, 2011.

[9] A. Albiol, L. Sanchis, A. Albiol, and J. M. Mossi, "Detection of parked vehicles using spatiotemporal maps," *IEEE Trans. on Intelligent Transportation Systems*, vol. 12, no. 4, pp. 1277-1291, 2011.

[10] Y. Tian, R. S. Feris, H. Liu, A. Hampapur, and M.-T. Sun, "Robust detection of abandoned and removed objects in complex surveillance videos," *IEEE Trans. on Systems, Man, and Cybernetics*, vol. 41, no. 5, pp. 565-576, 2011.

## Authors

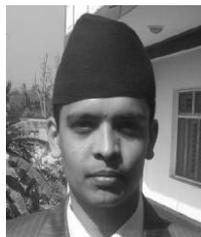

Deepak Ghimire

He received his B.E. degree in Computer Engineering from Pokhara University in 2017, M.S. degree in Computer Science and Engineering from Jeonbuk National University in 2011, and currently he is pursuing his Ph.D. degree. Research interest: image processing, computer vision, pattern recognition.

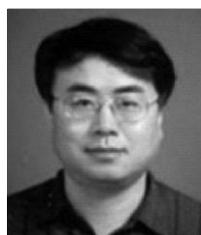

Joonwhoan Lee

He received his Bachelor degree in Electrical Engineering from Hanyang University in 1988, Master degree in Electrical and Electronics Engineering from KAIST University, Korea in 1982, and Ph.D. degree in Electrical and Computer Engineering from University of Missouri, USA in 1990. Research interests: image processing, audio processing, computer vision, emotion recognition.